\documentclass[final, 12pt]{alt2023} 

\title[deep Multi-Agent Reinforcement Learning with Hybrid Action Space]{A further exploration of deep Multi-Agent Reinforcement Learning with Hybrid Action Space}
\usepackage{times}
\usepackage{booktabs}

\altauthor{%
 \Name{Hongzhi Hua} \Email{hongzhihua@cqu.edu.cn}\\
 \addr College of Computer Science, Chongqing University, Shazheng
Street, Chongqing, 400044, China
 \AND
 \Name{Guixuan Wen} \Email{guixuanwen@cqu.edu.cn}\\
 \addr College of Computer Science, Chongqing University, Shazheng
Street, Chongqing, 400044, China
 \AND
 \Name{Kaigui Wu} \Email{kaiguiwu@cqu.edu.cn}\\
 \addr College of Computer Science, Chongqing University, Shazheng
Street, Chongqing, 400044, China
}

\begin{document}

\maketitle

\begin{abstract}%
  The research of extending deep reinforcement learning (drl) to multi-agent field has solved many complicated problems and made great achievements. However, almost all these studies only focus on discrete or continuous action space and there are few works having ever used multi-agent deep reinforcement learning to real-world environment problems which mostly have a hybrid action space. Therefore, in this paper, we propose two algorithms: deep multi-agent hybrid soft actor-critic (MAHSAC) and multi-agent hybrid deep deterministic policy gradients (MAHDDPG) to fill this gap. This two algorithms follow the centralized training and decentralized execution (CTDE) paradigm and could handle hybrid action space problems. Our experiences are running on multi-agent particle environment which is an easy multi-agent particle world, along with some basic simulated physics. The experimental results show that these algorithms have good performances.%
\end{abstract}


\section{Introduction}

In recent years, deep reinforcement learning (DRL)\citep{1,2} has been applied in many multi-agent fields to handle practical tasks, such as multiplayer games, autonomous driving, and the research of it has made great progress. This is crucial to building artificially intelligent systems that can effectively interact with humans and each other.

However, there are many problems and challenges about multi-agent reinforcement learning (MADRL)\citep{3,4}. Contrary to the theory, the realistic environment is changeable. In many settings, owing to restricted communication and partial observability, one agent can only realize the environmental changes, but ignores the affection of other agents' actions on its own action.

In previous researches on MADRL, researchers have made many explorations to improve the performance of  algorithms. The well-known MADDPG\citep{5} algorithm adopted the paradigm of centralized training and decentralized execution for the first time which is one of the most commonly used frameworks in MADRL field now. 

On the other hand, most popular multi-agent reinforcement learning algorithms ask the action space to be discrete or continuous only which is not consistent with the real world that the action space is usually discrete-continuous hybrid, such as real time strategic (RTS) games and robot movements. In this environment, each agent needs to select a discrete operation and its related continuous parameters at each timestep. In order to solve this problem, one approach is to  convert part of the continuous output from a fully continuous actor into discrete actions\citep{13}. Alternatively, one may use instead a fully discrete actor by discretizing the continuous actions, taking special care to prevent their number from exploding\citep{14}.

Then, a better solution is to learn directly in hybrid action space\citep{6}. Hybrid soft actor-critic (HSAC) is a successful algorithm to solve hybrid action space problem following this mind. Besides, there is a research which improved deep deterministic policy gradients (DDPG) algorithm to get good policy in hybrid action space. However, the attempt to apply those methods directly to multi-agent setting is not ideal because of instability in multi-agent environments.

In this work, we propose two novel approaches to address multi-agent problems in discrete-continuous hybrid action spaces with centralized training and decentralized execution framework: deep multi-agent hybrid soft actor-critic and multi-agent hybrid deep deterministic policy gradients algorithms. We extend HSAC to multi-agent settings and modify MADDPG to adopt to hybrid action space. Empirical results on multi-agent particle environment show the superior performance of our approaches compared to decentralized HSAC and hybrid DDPG in both cooperative and competitive environment.

\section{Background and related work}\label{sec2}

In this section, we discuss researches related to our proposed methods, including some attempts to extend the well-known single-agent DRL algorithm to the setting of hybrid action space.

\subsection{Deep Multi-agent Reinforcement Learning}\label{subsec2}

\begin{figure}[h]%
\centering
\includegraphics[width=0.7\textwidth]{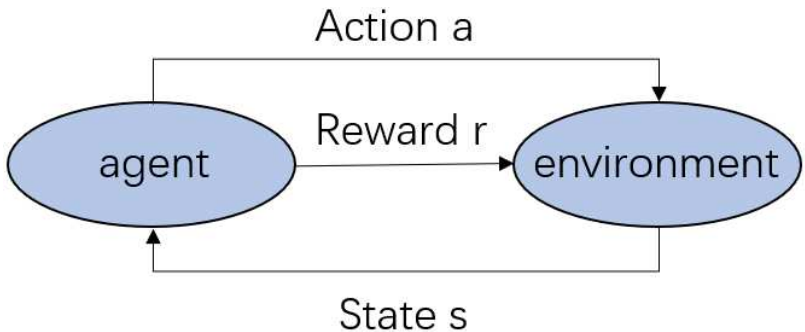}
\caption{the framework of deep reinforcement learning}\label{fig1}
\end{figure}

Drl has the abilities of feature extraction and sequence decision making which can solve many complex decision problems. It often uses Markov decision models to decompose problems and collects the state $s_t$, action $a_t$, reward $r_t$, the next state $s_{t+1}$ as a tuple ($s_t$ , $a_t$ , $r_t$ , $s_{t+1}$) at current time step to form a set ($S$,$A$,$R$,$S'$). The structure is shown in Figure \ref{fig1}. The optimization objective is strategy $\pi: s \rightarrow {a}$, and the cumulative reward which obtained at the time t by formula \ref{eq1} is maximized by optimizing it.

\begin{equation}
R=\sum\limits_{t'=t}^{T}y^{t'-t}r_t \label{eq1}
\end{equation}

Where $\gamma$ represents the attenuation factor.

Then, the Q function can be defined as $Q^\pi=E\left[R_t\mid s_t,a_t \right]$, and the optimal strategy $\pi^*$ is selected as the optimization goal to maximize the expectation of formula \ref{eq1}, which means that the formula \ref{eq2} is accurate.

\begin{equation}
Q^{\pi^*}\left(s,a\right)\geq Q^\pi\left(s,a\right) \forall s,a \in S,A \label{eq2}
\end{equation}

When deep reinforcement learning is applied to multi-agent systems, the environment becomes more complex, each agent needs to process more information, and the stability of the whole system will face greater challenges. If the learning method of single-agent is directly applied to multi-agents settings, it will be difficult to converge. To solve this problem, MADDPG algorithm proposed a new framework: centralized training and decentralized execution. 

\begin{figure}[h]%
\centering
\includegraphics[width=0.5\textwidth]{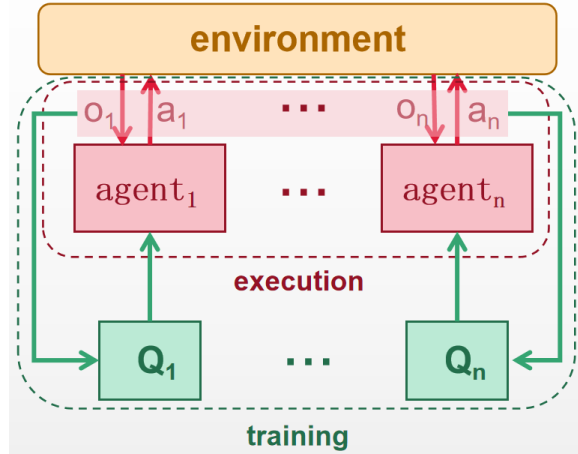}
\caption{the framework of centralized training and decentralized execution}\label{fig2}
\end{figure}

As Figure \ref{fig2} shows, MADDPG focuses on learning a centralized critic to take as input the actions of all agents and global state information during training for each agent. Besides, each agent interacts with the environment independently when taking as input the local observation and outputting the action value. Global information can promote the convergence of the network and improve the stability in the process of updating for training strategies with better performance. At the same time, the independent interaction between each agent and environment can effectively avoid dimensional disasters.

However, most of the existing multi-agent algorithms focus on only discrete action space or continuous action space which cannot be applied to many real-world environments. More effective algorithms for multi-agent problems with hybrid action space are still needed.

\subsection{Hybrid Soft Actor-Critic}\label{subsec3}

SAC\citep{7,8} is an exceptional off-policy algorithm that was originally proposed for continuous action space. This method aims to maximize both return of the agents and entropy of the actions which is significantly different from traditional algorithms which pay attention to maximize return of the agents. The main idea is to add an entropy bonus to the objective optimized by the agent, that is maximizing

\begin{equation}
E_\pi \left[\sum\nolimits_t \gamma^t \left(r_t+\alpha\mathcal{H}\left(\pi\left(\cdot\mid s_t\right)\right) \right)\right] \label{eq3}
\end{equation}

which can prevent the agent's strategy from converging prematurely to the local optimal solution. Where, policy $\pi$ is used to calculate distributions of the trajectory ($s_t,a_t$) from the strategy, that is, the entropy $\mathcal{H}$($\cdot$) of the strategy. $\alpha$ is used to pay more attention to entropy or reward, and a higher $\alpha$ will especially be conducive to explore possibility by encouraging agents to take actions which are more random. According to the soft policy iteration method, SAC updates the participant network by minimizing KL differences, and the following objectives need to be minimized:

\begin{equation}
J_\pi \left(\theta\right)=E_{s_t\sim D,a_t\sim\pi_\theta}\left[\alpha log\left(\pi_\theta\left(a_t\mid s_t\right)\right)-Q_\beta \left(s_t,a_t\right) \right]\label{eq4}
\end{equation}

Where, $\beta$ is the parameter of the critic network $Q_\beta$, $\theta$ is the parameter of the agent actor network $\pi_\theta$, and D represents the replay buffer. And it is failure to update the process by back propagating parameters because $a_t$ used in $Q_\beta(s_t,a_t)$ is obtained by sampling random policies $\pi_\theta$. Therefore, the reparameterization technique is practical for minimizing $J_\pi(\theta)$ by the random gradient. By adding random noise $\xi$ which follows the Gauss distribution, we can sample with a Gauss distribution. And then, the original sampling process is changed to: 

\begin{equation}
\Tilde{a}_\theta \left(s_t,\xi \right)=\tanh\left(\mu_\theta\left(s_t\right)+\sigma_\theta \left(s_t\right)\odot\xi \right)\label{eq5}
\end{equation}

Where, $\Tilde{a}_\theta \left(s_t,\xi \right)$ is the sample of probability distribution of agent's behavior $\pi_\theta\left(\cdot\mid s_t\right)$ , $\mu_\theta$ is mean value and $\sigma_\theta$ represents the standard deviation. Besides, $\tanh$ is used to ensure that the value of the agent's operation is limited within a certain range. By this approach, the optimization objectives are changed to:

\begin{equation}
\nabla_\theta J_\pi\left(\theta\right)=\nabla_\theta\alpha \log\left(\pi_\theta\left(a_t\mid s_t\right)\right)+\left(\nabla_{a_t}\alpha \log\left(\pi_\theta\left(a_t\mid s_t\right)\right)-\nabla_{a_t}Q\left(s_t,a_t\right)\right)\nabla_\theta\Tilde{a}_\theta \left(s_t,\xi \right) \label{eq6}
\end{equation}

And the critic network can be updated by minimizing the Berman error:

\begin{align}
J_Q\left(\beta\right)&=E_{s_t,a_t,r_t,s_{t+1}\sim D}\left[\frac{1}{2}\left(Q_\beta\left(s_t,a_t\right)-y\right)^2\right]\label{eq7}\\
y&=r\left(s_t,a_t\right)+\gamma E_{a_{t+1}\sim\pi_\theta\left(s_{t+1}\right)}\left[Q_{\overline{\beta}}\left(s_{t+1},a_{t+1}\right)-\alpha\log\left(\pi_\theta \left(a_{t+1}\mid s_{t+1}\right)\right)\right] \label{eq8}
\end{align}

To deal with the problem of discrete and continuous actions in SAC, there is a parameterization of the policy. One agent's operation $a$ can be represented by a combination of discrete actions $a^d= (a^d_1, . . . , a^d_D)$ and continuous parameters $a^c=(a^c_1, . . . , a^c_C)$. Each $a^d_i$ is an integer which represents the i-th discrete action that the agent may take. Each $a^c_j$ is an $m_j$ dimensional continuous vector which represents the j-th continuous action\citep{9,10}. So, actions are represented by tuples $(a^d_i,a^c_j)$. Assuming that the discrete actions are conditionally independent given the state s while the continuous parameters are conditionally independent given both s and the discrete actions. Then, yielding the following decomposition:

\begin{equation}
\begin{split}
\pi\left(a\mid s\right)&=\pi\left(a^d \mid s\right)\pi\left(a^c \mid s,a^d \right)\\
&=\prod_{i=1}^D \pi(a^d_i \mid s)\prod_{j=1}^C \pi(a^c_j \mid s,a^d)
\label{eq9}
\end{split}
\end{equation}

Here, the same letter $\pi$ is abused to denote both discrete probability mass functions and probability density functions applied to different components of the action.

\begin{figure}[h]%
\centering
\includegraphics[width=0.5\textwidth]{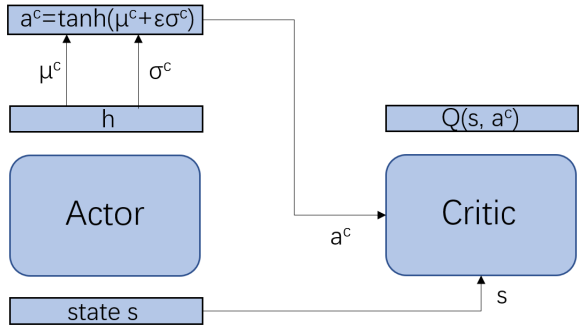}
\caption{architecture of standard SAC}\label{fig3}
\end{figure}

Figure \ref{fig3} shows the typical architecture of standard continuous SAC. By injecting standard normal noise $\xi$ and applying $\tanh$ nonlinearity to keep the action within a bounded range, actors can output the mean and standard deviation vectors $\mu^c$ and $\sigma^c$ which are used to sample action $a^c$. Then, the critic can estimate the corresponding Q-value by taking both state s and actor’s action $a^c$. 

\begin{figure}[h]%
\centering
\includegraphics[width=0.5\textwidth]{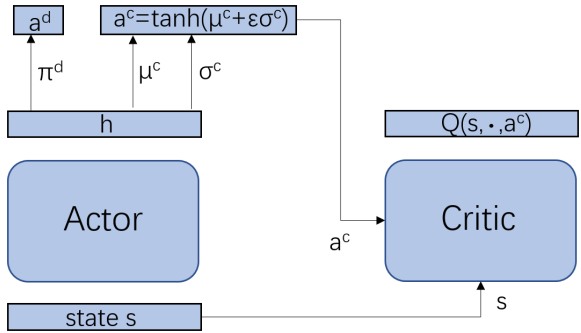}
\caption{architecture of HSAC}\label{fig4}
\end{figure}

SAC with hybrid action space requires a different policy parameterization which calls for a different network architecture. Therefore, Figure \ref{fig4} shows a situation where the agent must combine a discrete action $a^d$ with a set of independently sampled continuous parameters $a^c$. Here, a shared hidden state representation h produces additionally a discrete distribution $\pi^d$ to sample the discrete action $a^d$. Note that here, the value from critic’s output layer contains all discrete actions' predicted Q-values.

The SAC algorithm is based on the idea that the entropy additive value proportional to the entropy of $\pi(a\mid s)$ is given. As long as the action has a discrete part, the joint entropy definition with the weighted sum of discrete and continuous actions becomes:

\begin{equation}
\mathcal{H}(\pi(a^d,a^c\mid s))=\alpha^d \mathcal{H}(\pi(a^d\mid s))+\alpha^c \sum_{a^d}\pi(a^d\mid s)\mathcal{H}(\pi(a^c\mid a^d, s))
\label{eq10}
\end{equation}

Among them, the hyperparameters $\alpha^d$ and $\alpha^c$ encourage the exploration of discrete and continuous actions respectively. Besides, these two hyperparameters can be automatically adjusted in the learning process by using the same optimization techniques as soft actor-critic algorithms. Otherwise, they can be set to the same fixed value.

\subsection{Hybrid Deep Deterministic Policy Gradients}

DDPG is a classical algorithm which is usually used to deal with continuous action values. There are usually an actor network $\mu$, a critic network $Q$ and their target networks $\mu'$, $Q'$. The actor takes as input the state s and outputs the continuous action $a$ while the critic takes as input s, $a$ and outputs the Q-value $Q(s,a)$.

Compared with the standard temporal difference update originally used on Q-Learning \citep{11}, the update of critic network is basically unchanged, but the actor needs to provide the next-state action $a': \mu(s', \theta^\mu)$. Therefore, this process is greatly influenced by the actor's policy. 

For the actor, the goal is to minimize the difference between the current action $a$ and the optimal action in the same state. By a single backward pass over the critic network, it can provide gradients which indicate directions of change in action space. These gradients are placed at the output layer of actor network for its updating.

By employing the target network and replay memory D, the critic loss and actor update can be expressed by the following:

\begin{equation}
L_Q(\theta^Q)=E_{s_t,a_t,r_t,s_{t+1}\sim D}[(Q(s_t,a_t)-(r_t+\gamma Q'(s_{t+1},\mu'(s_{t+1}))))^2]
\label{eq16}
\end{equation}

\begin{equation}
\nabla_{\theta^\mu}\mu=E_{s_t\sim D}[\nabla_a Q(s_t,a\mid \theta^Q)\nabla_{\theta^\mu}\mu(s_t)\mid_{a=\mu(s_t)}]
\label{eq17}
\end{equation}

\begin{figure}[h]%
\centering
\includegraphics[width=0.5\textwidth]{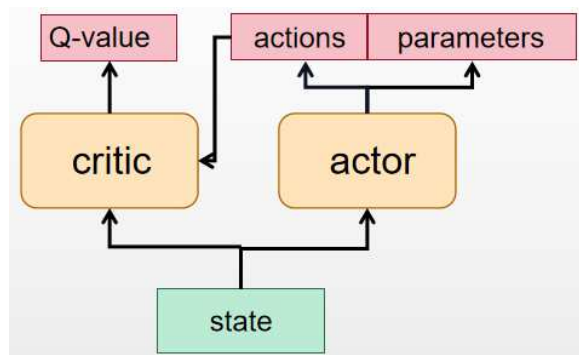}
\caption{architecture of HDDPG}\label{fig9}
\end{figure}

\citet{12} extended the DDPG algorithm into the hybrid action space. The definition of action space is as shown in section \ref{subsec3}. As Figure \ref{fig9} shows, at each timestep, the actor network takes as input the state which has two output layers: one for the discrete actions and another for the continuous parameters corresponding to these actions. The discrete action is chosen by be the maximally valued action output and paired with associated parameters from the parameter output. The critic network takes as input the discrete actions and continuous parameters for outputting a single scalar Q-value. So, the critic can provide gradients for discrete actions and continuous parameters while updating the actor.

\section{Method}\label{sec3}

To deal with hybrid action space problems in multi-agent environment, one attempt is to equip each agent with a decentralized algorithm in the independent learning paradigm which cannot get good results frequently in practice. The main factor for the failure is that each agent has a close relationship with other agents and these policy of them is ever-changing. Therefore, the environment is unstable or does not conform to Markov properties.

\subsection{Deep Multi-Agent Hybrid Soft Actor-Critic}\label{subsec6}

So, in this section, we extend HSAC to multi-agent setting based on CTDE training paradigm, and the network model is shown in Figure \ref{fig5}. We set up N agents in the environment to suit various situations, and the actor networks of these agents are a set: $\pi=\{\pi_1,\pi_2,...,\pi_n\}$, while the $Q^\beta_i$ represents the parameters of critic network $\beta_i$. For each agent $\pi_i$, it gets local observation $o_i$ from the environment as input and outputs the action $a_i$. In this paradigm, we can update the actor network of agent i by minimizing the target:

\begin{figure}[h]%
\centering
\includegraphics[width=0.5\textwidth]{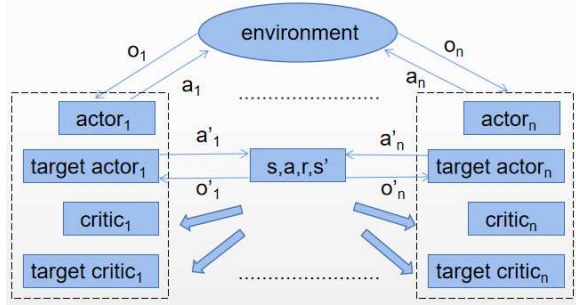}
\caption{The architecture of MAHSAC}\label{fig5}
\end{figure}

\begin{equation}
J_{\pi_i}(\theta_i)=E_{s\sim D,a\sim\pi_\theta}[\mathcal{H}(\pi(a_i\mid o_i))-Q_\beta (s,a)]
\label{eq11}
\end{equation}

Where, the joint entropy $\mathcal{H}(\pi(a_i\mid o_i))$ is defined as the formula \ref{eq10}. The state $s=\{o_1,o_2,...,o_n\}$ which is a set of agent i‘s local observation $o_i$ and the action $a=\{a_1,a_2,...,a_n\}$ which is a set of agent i‘s hybrid action $a_i$ are both stored in the experience replay buffer D. According to the theory of CTDE, the independent global critic network of each agent i should have taken the combination set of states and continuous actions from all agents as input. And the critic network takes the Q value corresponding to actions taken by agent i at timestep t as output to train strategy $\pi_i$. Especially, action $a$ taken as input by $Q_\beta$ is a set of continuous components $a_c$ from all agents. After the training, each agent only needs to observe its own local observation value $o_i$ to get a random strategy represented by the Gauss distribution. Besides, we can update agent i’s global critic network by minimizing the following:

\begin{equation}
J_Q (\beta_i)=E_{s,a,r,s'}[\frac{1}{2}(Q^{\beta_i}(s,a)-y)^2]
\label{eq12}
\end{equation}

Where, y is represented as:

\begin{equation}
y=r_i+\gamma E_{a'\sim \pi_{\overline{\theta}_i}}[Q^{\overline{\beta}_i}(s',a')-\mathcal{H}(\pi(a'_i\mid o'_i))]
\label{eq13}
\end{equation}

Here, $\gamma$ is the discount factor.

Our algorithm makes use of two soft Q-functions to mitigate positive bias in the policy improvement step that will degrade performance of value-based methods. We parameterize two soft Q-functions with parameters $\theta_i$, and calculate the Bellman error $J_{Q_i}$ of them separately. And the final $J_Q$ is just their average value. By this way, we can speed up training, especially on harder or complex tasks.

\subsection{Multi-Agent Hybrid Deep Deterministic Policy Gradients}

In this part, we also extend HDDPG to multi-agent setting based on CTDE paradigm. The structure of this algorithm is roughly the same as MAHSAC as the Figure \ref{fig5} shows. For each agent i, the actor $\mu_i$ takes as input the local observation $o_i$ and outputs the value of discrete actions and continuous parameters $a_i$. And we update $\mu_i$ by minimizing the following: 

\begin{equation}
L_\mu (\theta^{\mu_i})=E_{s_t,a_t\sim D}[-Q_i(s_t,a_t\mid \theta^{Q_i})]
\label{eq18}
\end{equation}

Where, the global state s and joint action $a$ whose definitions are same as section \ref{subsec6} are also stored in the experience replay buffer D for updating. Each agent i has its own independent global critic network which takes the combination set of states and actions as input. Here, the action $a$ taken by Q is a set of hybrid discrete actions and continue parameters from all agents. Besides, each agent's critic network will update by the loss:

\begin{equation}
L_Q(\theta^{Q_i})=E_{s_t,a_t,r_t,s_{t+1}\sim D}[(Q_i(s_t,a_t)-(r_t+\gamma Q'_i(s_{t+1},a_{t+1})))^2]
\label{eq19}
\end{equation}

It is very important to explore the action space, which helps agents maximize long-term rewards. We employ $\epsilon$-greedy exploration to hybrid action space where the discrete action is select randomly and the associated continuous parameters are sampled from a uniform random distribution with the probability $\epsilon$. And $\epsilon$ will change from 1.0 to 0.1 over the first 10,000 updates.

Compared with the traditional MADDPG algorithm, our algorithm modifies the structure of actor, so that it can output all the discrete actions and the corresponding continuous parameters at the same time, with changing the structure of critic according to this modification. Finally, we successfully adapt the MADDPG algorithm to the setting of hybrid action space, and have stronger practical value.

\subsection{Training Tricks}

In order to stabilize the training of deep neural network, each agent needs to add the actor target network $\pi_{\overline{\theta}_i}$ and critic target network $Q_{\overline{\beta}_i}$ for both two algorithms. Those target network parameters will be updated in soft mode every few episodes:

\begin{equation}
\overline{\theta}_i=\tau\theta_i+(1-\tau)\overline{\theta}_i\\
\label{eq14}
\end{equation}
\begin{equation}
\overline{\beta}_i=\tau\beta_i+(1-\tau)\overline{\beta}_i
\label{eq15}
\end{equation}

Obviously, the hyper-parameter $\tau$ can significantly affect the update of target network parameters.

In addition, we still use the technique of delaying the update. Every time actor updates N times, critic will update N times repeatedly in the same update cycle. The accuracy of critic's value function is largely influenced by the quality of the actor, since the actor determines the next-state action $a'$ in the update target. So, appropriate delay is helpful to shorten the training period and make critic network converge faster and better.

\section{Experiment}\label{sec4}

In our experiment, we adopt a simple multi-agent particle environment which consists of N agents and L landmarks inhabiting a two-dimensional world. We have modified this environment so that it can meet the demand of discrete-continuous hybrid action space. Agents may take physical actions and continuous parameters that get broadcasted to other agents in the environment. Each agent has accelerations on the X axis and Y axis respectively. The discrete actions is a set $(x^+,x^-,y^+,y^-)$ which means directions of acceleration to be applied. And the continuous parameters are values of acceleration which $\in$ [0,1] corresponding to discrete actions. There are some different scenarios with two different topics in the environment: cooperative and competitive. 

\subsection{Experimental Environment}\label{subsec4}

\begin{figure}[h]%
\centering
\includegraphics[width=0.5\textwidth]{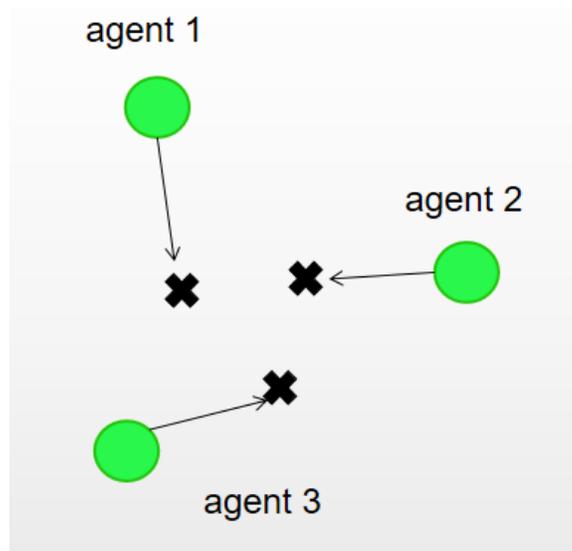}
\caption{Cooperative Navigation Scenario}\label{fig6}
\end{figure}

As shown in the Figure \ref{fig6}, there are three agents and three target points in a two-dimensional plane for the cooperative navigation scenario. Agents observe the relative positions of other agents and target points. Then, they will be collectively rewarded based on the proximity of any agent to each target. In addition, the agents occupy significant physical space and will be punished when they collide with each other. These agents need to learn to infer which targets they must cover and move there while avoiding other agents. 

\begin{figure}[h]%
\centering
\includegraphics[width=0.5\textwidth]{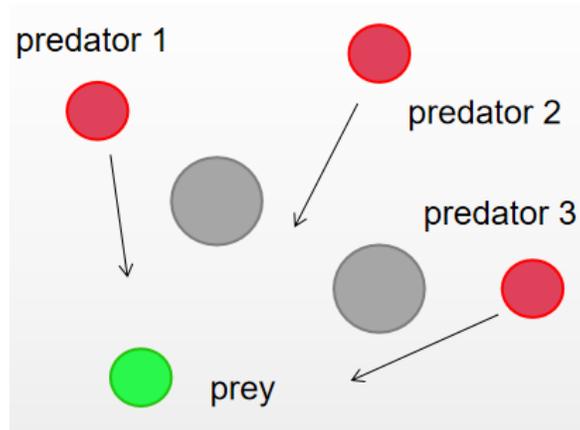}
\caption{Predator-Prey Scenario}\label{fig7}
\end{figure}

As we can see in Figure \ref{fig7}, there are 3 predators, 1 prey, and 2 obstacles in a two-dimensional plane for the Predator-Prey scenario. The goal of predators is to cooperate with each other to capture the prey, while the prey is aimed at avoiding the predators. Therefore, there is competition between the predators and the prey, and there is cooperation between the predators. 

\subsection{Experimental Result}\label{subsec5}

In the cooperative navigation scenario, agents will be rewarded according to the distance from the target and be punished when they collide with each other. Therefore, in this scenario, multi-agent and decentralization training models were compared in terms of the reward value, number of collisions and distance from the target.

\begin{figure}[h]%
\centering
\includegraphics[width=0.8\textwidth]{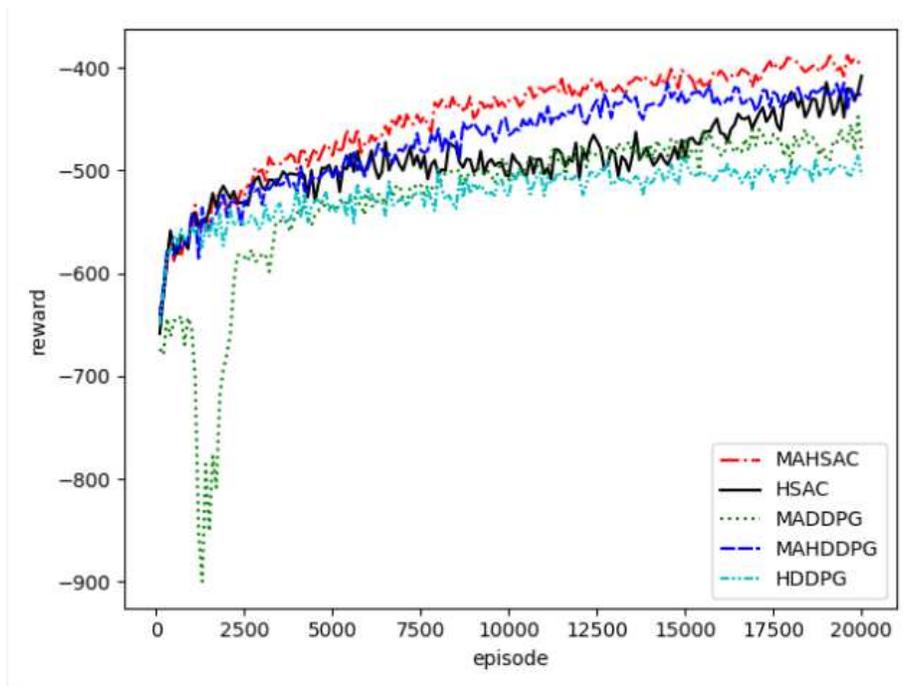}
\caption{agent reward on cooperative communication after 20000 episodes}\label{fig8}
\end{figure}

After 20000 episodes of training, the reward value curves of the agents are shown in Figure \ref{fig8}. We recorded the reward of the sum of all 3 agents per episode and save the average value every hundred episodes. Besides, we also applied MADDPG model to train in the original multi-agent particle environment as a comparison. As the fig shows, multi-agent algorithms significantly has a better performance than decentralization algorithms and MADDPG in training speed, reward and stability.

Table \ref{tab1} shows the number of collisions and distance from the target after algorithm convergence which corresponds to the performance test of agents. Compared with decentralized HSAC and hybrid DDPG, MAHSAC and MADDPG have lower values in these two respects which means theirs better performance.

\begin{table}[htbp]
\centering
\caption{\label{tab1}The average number of conflicts and the average proxy distance from landmarks per episode in the cooperative navigation scenario}
\begin{tabular}{lcl}
\toprule
agent & collisions  & dist \\
\midrule
MAHSAC    & 1.84395   & 0.242679  \\
HSAC    & 2.085   & 0.323587  \\
MAHDDPG    & 1.7633   & 0.269789  \\
HDDPG    & 1.8143   & 0.379369  \\
\bottomrule
\end{tabular}
\end{table}

\begin{table}[h]
\centering
\caption{\label{tab2}Average number of prey touches by predator per episode on the predator-prey scenario}
\begin{tabular}{lcl}
\toprule
agent & adversary & touches  \\
\midrule
MAHSAC    & MAHSAC   & 2.89785  \\
MAHSAC    & HSAC   & 20.587596  \\
HSAC    & MAHSAC   & 2.2201  \\
HSAC    & HSAC   & 1.8726  \\
MAHDDPG    & MAHDDPG   & 0.7314  \\
MAHDDPG    & HDDPG   & 2.3413  \\
HDDPG    & MAHDDPG   & 1.3010  \\
HDDPG    & HDDPG   & 1.4565  \\
\bottomrule
\end{tabular}
\end{table}

The above data shows that our methods can train policy that will achieve the goal faster and better in the cooperative environment than decentralization algorithms. This proves the advantages of the joint hybrid strategy between explicit coordination agents. Besides, MAHSAC outperforms MAHDDPG after convergence. We believe that this is due to SAC's maximum entropy mechanism which can effectively prevent the agent's strategy from converging to the local optimal solution prematurely.

And in the Predator-Prey scenario, due to the competition between predators and prey, the reward value curve of agents is unstable and cannot reflect the real performance of algorithms. Therefore, we recorded the average number of times prey was touched by predator per episode. Here, in order to concretely show the differences of performance between different algorithms, we set four situations where multi-agent setting and decentralization methods were respectively applied to predator and prey agents.

The test results are shown as Table \ref{tab2}, a higher number of collisions indicates that the predators can catch their prey earlier. As we can see, multi-agent predators are far more successful at chasing decentralized prey than the converse and decentralized agent had a bad performance when directly pitted against multi-agent setting policy in all cases. This demonstrates that our approaches reflect stronger learning ability and convergence performance in the fight against decentralized agents with the competitive scenario.

\section{Conclusion}\label{sec5}

This paper makes an attempt to extend HSAC and hybrid DDPG to multi-agent settings while providing two novel ways to apply deep reinforcement learning in multi-agent environments to handle practical problems with discrete-continuous hybrid action space which further fills the vacancy in this area. Under the paradigm of centralized training and decentralized execution, we propose MAHSAC,MAHDDPG algorithms and the experimental results show their superiority to decentralized hybrid methods under an easy multi-agent particle world with basic simulated physics. For future work, we will apply some advanced and effective neural network structures to the research of muiti-agent hybrid action space setting. We hope that this will make our algorithms to adapt to more diversified and complicated environments while showing more excellent performance.



\acks{This paper is the result of Project No.2022CDJKYJH023 supported by the Fundamental Research Funds for the Central Universitie.}

\bibliography{alt2023-sample}

\end{document}